\pdfoutput=1

\documentclass[11pt]{article}

\usepackage[final]{acl}

\usepackage{times}
\usepackage{latexsym}
\usepackage[T1]{fontenc}
\usepackage[utf8]{inputenc}
\usepackage{microtype}
\usepackage{inconsolata}
\usepackage{graphicx}

\usepackage{todonotes}
\usepackage{booktabs}
\usepackage{array}
\usepackage{amssymb}
\usepackage{tabularx}
\usepackage{adjustbox}

\usepackage{booktabs}
\usepackage{siunitx}
\usepackage{multirow}
\usepackage{rotating}
\usepackage{caption}
\usepackage{threeparttable}
\usepackage{array}
\usepackage{graphicx}
\usepackage{placeins}

\title{data2lang2vec: Data Driven Typological Features Completion}

\author{
    \textbf{Hamidreza Amirzadeh}$^{1,2}$ \ \textbf{Sadegh Jafari}$^{1,3}$ \ \textbf{Anika Harju}$^1$ \\
    \textbf{Rob van der Goot}$^1$ \\
    $^1$ IT University of Copenhagen, Denmark \\
    $^2$ Sharif University of Technology, Iran \\
    $^3$ Iran University of Science and Technology, Iran \\
    \normalsize \texttt{hamid.amirzadeh78@sharif.edu}, \ \texttt{sadegh\_jafari@comp.iust.ac.ir} \\
    \normalsize \texttt{\{aniha, robv\}@itu.dk}
}

\begin{document}
\maketitle
\begin{abstract}
Language typology databases enhance multi-lingual Natural Language Processing (NLP) by improving model adaptability to diverse linguistic structures. The widely-used lang2vec toolkit integrates several such databases, but its coverage remains limited at 28.9\%.
Previous work on automatically increasing coverage predicts missing values based on features from other languages or focuses on single features, we propose to use textual data for better-informed feature prediction.
To this end, we introduce a multi-lingual Part-of-Speech (POS) tagger, achieving over 70\% accuracy across 1,749 languages, and experiment with external statistical features and a variety of machine learning algorithms. 
We also introduce a more realistic evaluation setup, focusing on likely to be missing typology features, and show that our approach outperforms previous work in both setups.
\end{abstract}

\section{Introduction}
Language typology databases contain information about specific languages, for example subject-verb order. These databases are not only used to document and study languages~\cite{yu-etal-2021-language}, and their relations to each other~\cite{toossi2024reproducibility}, but have also shown to be beneficial for multi-lingual Natural Language Processing (NLP) applications. By informing the model explicitly about the differing structures to expect, the model can more easily adapt to other, even unseen languages~\cite{ustun-etal-2022-udapter}.

In NLP, \textit{lang2vec}~\cite{littell-etal-2017-uriel} toolkit is commonly used, probably because of its ease of use and the coverage of languages. \textit{lang2vec} includes a collection of previously existing databases, namely: WALS~\cite{vastl-etal-2020-predicting} PHOIBLE~\cite{dediu2016defining} Ethnologue~\cite{ethnologue}, and Glottolog~\cite{glottolog}, which are all converted to a uniform format, where each feature is represented as a binary feature. In total, it includes 4005 languages and 289 features, However, even after combining multiple sources, coverage is still only 28.9\%.


In \textit{lang2vec}, KNN is used to obtain values for all feature-language combinations, based on the hypothesis that languages that are similar to each other in many features will also be similar to each other for unknown features. They report an accuracy of 92.93 in a 10-fold setup with all included features. However, this solution leads to similar languages becoming more similar to each other and more distant languages more different. Hence, other works have made attempts to predict smaller sets of features based on texts from the target languages~\cite{barbieri-etal-2022-xlm}, often combined with automated syntactic analyses~\cite{he-sagae-2019-syntactic}. However, a comprehensive analysis of a more complete feature set is missing. Hence, we contribute:
\begin{itemize}
\setlength{\itemsep}{0pt}
    \item We propose a more realistic evaluation framework for typological feature prediction, focusing on identifying feature values that have gold-standard annotations but are likely to have been missing in \textit{lang2vec}.
    \item We evaluate text-based approaches to complete the whole inventory of \textit{lang2vec} features as well as statistical features about the languages. We show that only certain features benefit from the POS tags, and statistical features are more informative. 
    \item We also provide a multi-lingual POS tagger with an estimated performance of $>70$ accuracy for 1,749 languages, completed \textit{lang2vec} data, and a toolkit to provide meta-data for languages.
    \footnote{Our code is freely available at \url{https://github.com/hamid-amir/data_lang2vec}}
\end{itemize}


\section{Data and Methodolgy}
\label{sec:models}

\subsection{Features}
\label{sec:feats}
We use three groups of features for the prediction, each described in a paragraph below:

\paragraph{\textit{lang2vec} features}
Features directly extracted from the \textit{lang2vec} database:
\begin{itemize}
\setlength{\itemsep}{0pt}
\item phylogeny: Use the \textit{fam} feature from \textit{lang2vec}. It has 3719 dimensions.
\item lang\_id: ISO 639-3 code is a unique identifier for each language.
\item feat\_id: The \textit{lang2vec} identifier of the feature.
\end{itemize}

\paragraph{External features}
We use the following continuous features:
\begin{itemize}
\setlength{\itemsep}{0pt}
\item lang\_fam: The language family to which a language belongs. We use the main families of Glottolog for each language.
\item geo\_lat: latitude location of language, taken from Glottolog 5.0.
\item geo\_long: Longitude location of language, taken from Glottolog 5.0.
\item wiki\_size: The Wikipedia size as reported by Wikipedia\footnote{\url{https://en.wikipedia.org/wiki/List_of_Wikipedias}}.
\item num\_speakers: Taken from the ASJP database~\cite{ASJP}.
\end{itemize}

Furthermore, we add the following n-hot features:
\begin{itemize}
\setlength{\itemsep}{0pt}
\item aes\_status: The Agglomerated Endangerment Status (AES) scale is derived from data provided by Glottolog 5.0~\cite{glottolog}, which, in turn, sourced its data from ELCat~\cite{elcat},  UNESCO Atlas of the World's Languages in Danger~\cite{moseley2010atlas}, and Ethnologue~\cite{ethnologue}. It has 6 possible values.
\item lang\_group: ~\citet{joshi-etal-2020-state} propose a taxonomy based on the number of language resources for languages, they identify 6 groups.
\item scripts: since a language might be written in multiple scripts, we support this as an n-hot feature list. We take the information from ~\citet{kargaran-etal-2024-glotscript-resource}, and remove the Brai script (Braille), as it is annotated inconsistently.
\item feat\_name: The features in \textit{lang2vec} have short, sometimes overlapping names (e.g., S\_ADPOSITION\_BEFORE\_NOUN). We split these names into word unigrams (by `\_') and use them as binary features.
\end{itemize}

\paragraph{Textual features}
We choose to use the LTI LangID corpus~\cite{brown-2014-non} version 5 as a source for our textual data, as it has the widest language coverage to the best of our knowledge. 
We use the official mapping of retired ISO 639-3 codes and remove all texts that have an invalid iso639-3 language code as well as macro-languages. We end up with data for 2,134 languages (note that there is a total of 7,077 languages in ISO 639-3, of which approximately half is estimated to have a standard written form).

\label{sec:pos_tagger}
Text-based data is unsuitable for our classifier because of the large amount of features and lack of overlap across the languages, resulting in poor performance. Hence, we experiment with POS tags as features.
We trained POS taggers using various multi-lingual models, including mBERT~\cite{devlin-etal-2019-bert}, twitter-XLM-roBERTa~\cite{barbieri-etal-2022-xlm}, InfoXLM~\cite{chi-etal-2021-infoxlm}, mDeBERTa-v3~\cite{he2020deberta}, XLM-roBERTa~\cite{conneau-etal-2020-unsupervised}, mLUKE~\cite{yamada-etal-2020-luke}, and TwHIN-BERT~\cite{el2022twhin}. The models were trained on all UD V2.14 training splits (up to 200,000 words per treebank) using MaChAmp~\cite{van-der-goot-etal-2021-massive}. We jointly trained tokenization and POS tagging, evaluating on 150 languages. InfoXLM-large~\cite{chi-etal-2021-infoxlm} performed best on the 70 unseen languages. This model was further trained on full UD treebanks for final predictions (results in Appendix~\ref{app:posresults}).

\subsection{Models}
We train models using KNN, Logistic Regression, Gradient Boosting, Decision Trees, and Random Forests. For efficiency reasons, we first evaluated all classifiers on a small sample, and based on this focused mainly on KNN (baseline) and Random Forest (best performance).


\begin{figure}[t]
\centering
\centering
\includegraphics[width=1.\linewidth]{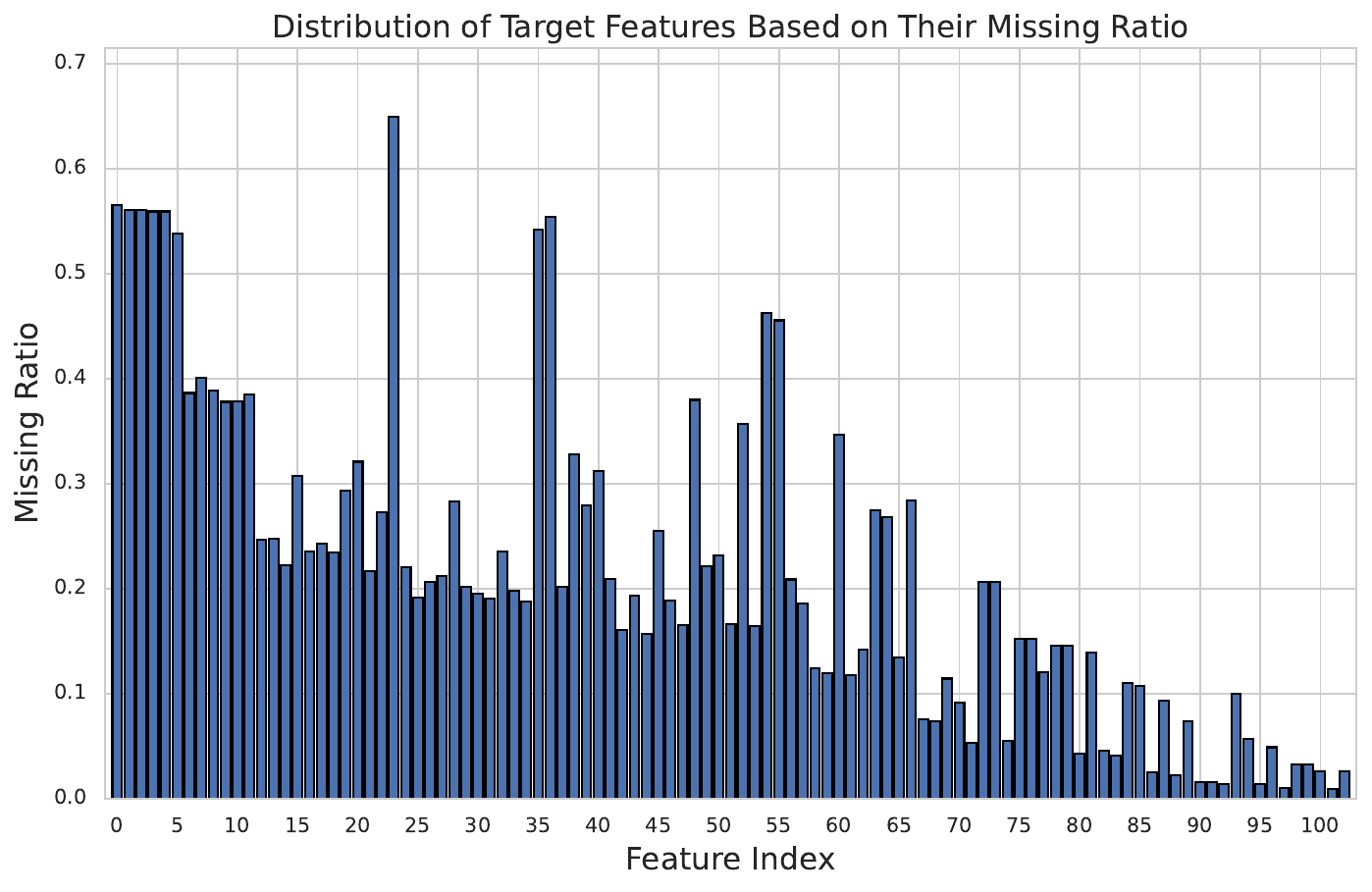}
\hspace{0.02\textwidth}
\caption{Missing ratio distribution of our target features. Higher bars indicate that feature has more probable missing values, and vice versa. Features with a missing ratio above 0.5 are listed in Table \ref{tab:second_eval}.
}
\label{fig:feat_dist}
\end{figure}

\section{Setup}

We choose to use the syntax and phonology features of WALS as our main focus.
In addition to performing a (random) k-fold split, a classifier is constructed to predict which features are likely to be missing, a task we refer to as ``feature presence classification''.
This way, we avoid overestimating our performance (e.g. predicting whether English has
Subject-verb-object order is easier compared to most other features).

\subsection{Feature Presence Classification}
\label{sec:findMissing}
To train a predictor for identifying ``likely missing values'' a binary classifier is employed to predict whether a target feature for a specific language is present in \textit{lang2vec} or not. We use the same features as we use for our prediction model (Section~\ref{sec:feats}), except for the text-based features. We experimented with several common machine learning classifiers, and additionally applied hyperparameter optimization using the Optuna framework ~\cite{optuna_2019}.
After obtaining the best model, a dataset is created by ranking the present features based on the model's confidence to estimate the final target feature predictions.
We then used top 20\% that was most likely to be missing for evaluation purposes (see also Appendix~\ref{app:setup} for an overview of our setup).


\subsection{Typological Features Classification}
There are 125 features related to syntax and phonology in WALS that we aim to predict. We developed a distinct classifier for each of these 125 features. This approach was deliberately chosen because the designed and prepared features may not uniformly contribute to the prediction accuracy across different features. By employing distinct classifiers, we prevent weight sharing, which enhances the prediction performance for each specific feature.

The same feature set used in the "feature presence classifier" was employed, with the addition of POS tags. We use n-gram counts with n in the range of 3-5, resulting in approximately one million dimensions. To mitigate the impact of this high dimensionality, which could overshadow features with fewer dimensions, Principal Component Analysis (PCA) was applied for dimensionality reduction, introducing a hyperparameter to specify the number of dimensions to retain.
We use Optuna~\cite{optuna_2019} for hyperparameter optimization and feature selection.

\begin{table*}
    \centering
    \footnotesize 
    \setlength{\tabcolsep}{3pt} 
    \resizebox{1.0\linewidth}{!}{
    \begin{tabular}{l S[table-format=2.2] S[table-format=2.2] *{15}{c}}
        \toprule
        \multirow{2}{*}{\textbf{Target Feature}} & 
        \multicolumn{2}{c}{\textbf{F1 Score (\%)}} & 
        \multicolumn{15}{c}{\textbf{Feature Selection}} \\
        \cmidrule(lr){2-3} \cmidrule(l){4-18}
        & {\textbf{KNN}} & {\textbf{Ours}} & 
        \rotatebox{90}{\textbf{lang\_id}} & 
        \rotatebox{90}{\textbf{feat\_id}} & 
        \rotatebox{90}{\textbf{geo\_lat}} & 
        \rotatebox{90}{\textbf{geo\_long}} & 
        \rotatebox{90}{\textbf{lang\_group}} & 
        \rotatebox{90}{\textbf{aes\_status}} & 
        \rotatebox{90}{\textbf{wiki\_size}} & 
        \rotatebox{90}{\textbf{num\_speakers}} & 
        \rotatebox{90}{\textbf{lang\_fam}} & 
        \rotatebox{90}{\textbf{scripts}} & 
        \rotatebox{90}{\textbf{feat\_name}} & 
        \rotatebox{90}{\textbf{phylogeny}} &
        \rotatebox{90}{\textbf{phylo\_n\_comp}} &
        \rotatebox{90}{\textbf{LTI\_LangID}} &
        \rotatebox{90}{\textbf{LTI\_LangID\_n\_comp}} \\
        \midrule
        S\_VOX & 5.71 & \textbf{69.76} & $\times$ & \checkmark & \checkmark & \checkmark & \checkmark & \checkmark & $\times$ & $\times$ & $\times$ & $\times$ & $\times$ & $\times$ & $\times$ & $\times$ & $\times$ \\
        \midrule
        S\_OBLIQUE\_AFTER\_VERB & 11.76 & \textbf{68.57} & $\times$  & \checkmark & $\times$ & \checkmark & \checkmark & $\times$ & $\times$ & $\times$ & $\times$ & \checkmark & \checkmark & \checkmark & 67 & $\times$ & $\times$ \\
        \midrule
        S\_POSSESSIVE\_DEPMARK & 56.86 & \textbf{69.39} & \checkmark  & \checkmark & \checkmark & $\times$ & $\times$ & $\times$ & $\times$ & \checkmark & \checkmark & \checkmark & \checkmark & \checkmark & 81 & \checkmark & 417 \\
        \midrule
        \midrule
        \textbf{Average performance} & 77.43 & \textbf{83.05} & 0.42 & 0.46 & 0.60 & 0.57 & 0.47 & 0.56 & 0.57 & 0.41 & 0.44 & 0.51 & 0.48 & 0.75 & 56 & 0.23 & 414 \\
        \bottomrule
    \end{tabular}}
    \caption{
    K-fold cross-validation results for both KNN and our method.  The presence of each of our curated features for predicting each target feature is indicated by \checkmark{} or $\times$ symbols, obtained by Optuna. We highlight the 3 target features with the largest performance improvements over KNN from the original \textit{lang2vec} paper. The last row shows feature usage percentages; for instance, \texttt{LTI\_LangID} was used in 23\% of the target features. \texttt{phylo\_n\_comp} and \texttt{LTI\_LangID\_n\_comp} are PCA hyperparameters for \texttt{phylogeny} and \texttt{LTI\_LangID}.
    }
    \label{tab:first_eval}
\end{table*}

\begin{table}
    \centering
    \footnotesize 
    \setlength{\tabcolsep}{2pt} 
    \resizebox{0.9\linewidth}{!}{
    \begin{tabular}{l S[table-format=3.2] S[table-format=3.2] S[table-format=3.2]}
        \toprule
        \multirow{2}{*}[-0.5ex]{\textbf{Target Feature}} & 
        \textbf{Missing Ratio} & 
        \multicolumn{2}{c}{\textbf{F1 Score (\%)}} \\
        \cmidrule(lr){3-4}
        & & {KNN} & {Ours} \\
        \midrule
        S\_COM\_VS\_INST\_MARK & 0.65 & 30.18 & \textbf{42.42} \\
        S\_SVO & 0.56 & 80.00 & \textbf{83.62} \\
        S\_VSO & 0.56 & 94.96 & \textbf{96.21} \\
        S\_OSV & 0.56 & 99.85 & \textbf{100.0} \\
        S\_SOV & 0.55 & 82.71 & \textbf{86.98} \\
        S\_OVS & 0.55 & 99.27 & \textbf{99.42} \\
        S\_ANY\_REDUP & 0.55 & 52.63 & 52.63 \\
        S\_NUMCLASS\_MARK & 0.54 & 67.69 & \textbf{73.76} \\
        S\_VOS & 0.53 & 98.79 & 98.79 \\
        \midrule
        \midrule
        \textbf{Avgerage performance} & \  & 74.78 & \textbf{76.91} \\
        \bottomrule
    \end{tabular}
    }
    \caption{Results of our method and the KNN baseline for the nine target features with the highest likelihood of being missing in our proposed evaluation setup.}
    \label{tab:second_eval}
\end{table}

\section{Results and Evaluations}
\subsection{Feature Presence} 
For the feature presence classifier, we initially created a sub-sample by selecting 300 random languages to extract data from. Subsequently, all classifiers (Section~\ref{sec:models}) were trained and optimized to evaluate their performance in predicting the probability of whether a feature and language combination is missing in \textit{lang2vec}. Ultimately, the best-performing classifier, Gradient Boosting, was trained on all languages using the previously optimized settings, achieving the highest F1 score of 98.61. We then used this model to rank typological features based on the model's confidence in classifying them as missing, selecting the top 20\% for evaluation purposes. 
We introduce the ``Missing Ratio'', which is calculated by dividing number of language features in a target feature determined to be missing (i.e. it is among top 20\% probable missings) by the total number of languages in that target feature. Figure~\ref{fig:feat_dist} plots the missing ratio distribution, showing a large disparity in the likelihood that certain features are missing.
Additional results from other models and details on hyperparameter optimization can be found in Appendix \ref{app:finding-tuning}.

\subsection{Typological Features Results}

\paragraph{Results of k-fold}
In the original \textit{lang2vec} study, missing typological features were imputed using a simple KNN classifier, while the present features remained unchanged. To establish a baseline for our work, we applied the same KNN approach to predict the values of the 125 target features that were already present, enabling us to evaluate the accuracy of this method and compare it to ours. Table \ref{tab:first_eval} displays several of our significant gains compared to the KNN method for predicting target features across all existing languages. Additionally, this table highlights the importance of each of our introduced features in predicting each target feature. 
The most effective feature is 'pylogency,' useful for predicting 75\% of target features, while POS tag features from LTI\_langID are the least effective, contributing to only 23\%. This aligns with the fact that language family is more informative, while POS tags feature are mainly useful for predicting word order. Since none of our curated features relate to phonetics, we cannot expect to do better in the related target features. Future work could introduce phonetic features to improve results.

\paragraph{Results of likely missing values}
For a more realistic evaluation, we employed the same methodology but focused exclusively on the top 20\% of the missing features identified in Section \ref{sec:findMissing}, rather than considering all present values. Table \ref{tab:second_eval} presents the performance of both the KNN method and our approach for predicting target features with a missing ratio above 0.5 (see Figure \ref{fig:feat_dist}). There are nine target features with a missing ratio exceeding 0.5, and for all of these critical features, our method either outperforms or performs on-par with the KNN approach.

\section{Conclusion}

In this paper, we focused on predicting missing values for syntax and phonology features in the \textit{lang2vec} database. Besides the commonly used features from the \textit{lang2vec} database itself, we experiment with statistical features of the languages and POS tags obtained from textual data. 
We showed that the features from \textit{lang2vec} are moderately useful, but the external statistical features are most beneficial. The POS tagging features are only useful for selected features. 
We also provide a more realistic evaluation setting compared to previous work, which used k-fold; we propose to focus our evaluation metrics on features that are likely to be missing.
Our proposed model with all features outperforms the KNN-based approach of \textit{lang2vec} in both setups, especially for features with higher probabilities of missing values.

\section{Limitations}
We focused on WALS for predicting target features, though the same approach could be applied to other typological resources in \textit{lang2vec} (SSWL and Ethnologue) or outside of \textit{lang2vec}, for example, GramBank~\cite{haynie-etal-2023-grambanks}.  
Moreover, models were trained and optimized on a small subset of languages before applying the best one to the full dataset due to computational constraints. Finally, we focused on a subset of the world’s languages and used iso639-3 as the definitive label, acknowledging its limitations~\cite{morey2013language}.

\section{Acknowledgements}
We would like to thank Esther Ploeger for providing relevant references and initial ideas. Christian Hardmeier for his thoughts on POS tagging evaluation, and Lottie Rosamund Greenwood for maintaining the HPC at ITU. 
We also extend our gratitude to the \textit{Speech and Language Processing Lab}, led by Professor Hosein Sameti at Sharif University of Technology, for providing part of the computational resources necessary for this work.

\bibliography{papers}

\begin{thebibliography}{29}
\expandafter\ifx\csname natexlab\endcsname\relax\def\natexlab#1{#1}\fi

\bibitem[{Akiba et~al.(2019)Akiba, Sano, Yanase, Ohta, and Koyama}]{optuna_2019}
Takuya Akiba, Shotaro Sano, Toshihiko Yanase, Takeru Ohta, and Masanori Koyama. 2019.
\newblock Optuna: A next-generation hyperparameter optimization framework.
\newblock In \emph{Proceedings of the 25th {ACM} {SIGKDD} International Conference on Knowledge Discovery and Data Mining}.

\bibitem[{Barbieri et~al.(2022)Barbieri, Espinosa~Anke, and Camacho-Collados}]{barbieri-etal-2022-xlm}
Francesco Barbieri, Luis Espinosa~Anke, and Jose Camacho-Collados. 2022.
\newblock \href {https://aclanthology.org/2022.lrec-1.27} {{XLM}-{T}: Multilingual language models in {T}witter for sentiment analysis and beyond}.
\newblock In \emph{Proceedings of the Thirteenth Language Resources and Evaluation Conference}, pages 258--266, Marseille, France. European Language Resources Association.

\bibitem[{Brown(2014)}]{brown-2014-non}
Ralf Brown. 2014.
\newblock \href {https://doi.org/10.3115/v1/D14-1069} {Non-linear mapping for improved identification of 1300+ languages}.
\newblock In \emph{Proceedings of the 2014 Conference on Empirical Methods in Natural Language Processing ({EMNLP})}, pages 627--632, Doha, Qatar. Association for Computational Linguistics.

\bibitem[{Chi et~al.(2021)Chi, Dong, Wei, Yang, Singhal, Wang, Song, Mao, Huang, and Zhou}]{chi-etal-2021-infoxlm}
Zewen Chi, Li~Dong, Furu Wei, Nan Yang, Saksham Singhal, Wenhui Wang, Xia Song, Xian-Ling Mao, Heyan Huang, and Ming Zhou. 2021.
\newblock \href {https://doi.org/10.18653/v1/2021.naacl-main.280} {{I}nfo{XLM}: An information-theoretic framework for cross-lingual language model pre-training}.
\newblock In \emph{Proceedings of the 2021 Conference of the North American Chapter of the Association for Computational Linguistics: Human Language Technologies}, pages 3576--3588, Online. Association for Computational Linguistics.

\bibitem[{Conneau et~al.(2020)Conneau, Khandelwal, Goyal, Chaudhary, Wenzek, Guzm{\'a}n, Grave, Ott, Zettlemoyer, and Stoyanov}]{conneau-etal-2020-unsupervised}
Alexis Conneau, Kartikay Khandelwal, Naman Goyal, Vishrav Chaudhary, Guillaume Wenzek, Francisco Guzm{\'a}n, Edouard Grave, Myle Ott, Luke Zettlemoyer, and Veselin Stoyanov. 2020.
\newblock \href {https://doi.org/10.18653/v1/2020.acl-main.747} {Unsupervised cross-lingual representation learning at scale}.
\newblock In \emph{Proceedings of the 58th Annual Meeting of the Association for Computational Linguistics}, pages 8440--8451, Online. Association for Computational Linguistics.

\bibitem[{Dediu and Moisik(2016)}]{dediu2016defining}
Dan Dediu and Scott Moisik. 2016.
\newblock Defining and counting phonological classes in cross-linguistic segment databases.
\newblock In \emph{LREC 2016: 10th International Conference on Language Resources and Evaluation}, pages 1955--1962. European Language Resources Association (ELRA).

\bibitem[{Devlin et~al.(2019)Devlin, Chang, Lee, and Toutanova}]{devlin-etal-2019-bert}
Jacob Devlin, Ming-Wei Chang, Kenton Lee, and Kristina Toutanova. 2019.
\newblock \href {https://doi.org/10.18653/v1/N19-1423} {{BERT}: Pre-training of deep bidirectional transformers for language understanding}.
\newblock In \emph{Proceedings of the 2019 Conference of the North {A}merican Chapter of the Association for Computational Linguistics: Human Language Technologies, Volume 1 (Long and Short Papers)}, pages 4171--4186, Minneapolis, Minnesota. Association for Computational Linguistics.

\bibitem[{Eberhard et~al.(2024)Eberhard, Simons, and Fennig}]{ethnologue}
David~M. Eberhard, Gary~F. Simons, and Charles~D. Fennig. 2024.
\newblock \href {http://www.ethnologue.com.} {Ethnologue: Languages of the world. twenty-seventh edition.}

\bibitem[{El-Kishky et~al.(2022)El-Kishky, Markovich, Park, Verma, Kim, Eskander, Malkov, Portman, Samaniego, Xiao et~al.}]{el2022twhin}
Ahmed El-Kishky, Thomas Markovich, Serim Park, Chetan Verma, Baekjin Kim, Ramy Eskander, Yury Malkov, Frank Portman, Sof{\'\i}a Samaniego, Ying Xiao, et~al. 2022.
\newblock Twhin: Embedding the twitter heterogeneous information network for personalized recommendation.
\newblock In \emph{Proceedings of the 28th ACM SIGKDD conference on knowledge discovery and data mining}, pages 2842--2850.

\bibitem[{Hammarström et~al.(2024)Hammarström, Forkel, Haspelmath, and Bank}]{glottolog}
Harald Hammarström, Robert Forkel, Martin Haspelmath, and Sebastian Bank. 2024.
\newblock \href {https://doi.org/10.5281/zenodo.10804357} {Glottolog 5.0.}

\bibitem[{Haynie et~al.(2023)Haynie, Blasi, Skirg{\aa}rd, Greenhill, Atkinson, and Gray}]{haynie-etal-2023-grambanks}
Hannah~J. Haynie, Dami{\'a}n Blasi, Hedvig Skirg{\aa}rd, Simon~J. Greenhill, Quentin~D. Atkinson, and Russell~D. Gray. 2023.
\newblock \href {https://doi.org/10.18653/v1/2023.sigtyp-1.17} {Grambank{'}s typological advances support computational research on diverse languages}.
\newblock In \emph{Proceedings of the 5th Workshop on Research in Computational Linguistic Typology and Multilingual NLP}, pages 147--149, Dubrovnik, Croatia. Association for Computational Linguistics.

\bibitem[{He et~al.(2020)He, Liu, Gao, and Chen}]{he2020deberta}
Pengcheng He, Xiaodong Liu, Jianfeng Gao, and Weizhu Chen. 2020.
\newblock Deberta: Decoding-enhanced bert with disentangled attention.
\newblock In \emph{International Conference on Learning Representations}.

\bibitem[{He and Sagae(2019)}]{he-sagae-2019-syntactic}
Taiqi He and Kenji Sagae. 2019.
\newblock \href {https://typology-and-nlp.github.io/2019/assets/2019/papers/7.pdf} {Syntactic typology from plain text using language embeddings}.
\newblock In \emph{Proceedings of the First Workshop on Typology for Polyglot NLP}.

\bibitem[{Joshi et~al.(2020)Joshi, Santy, Budhiraja, Bali, and Choudhury}]{joshi-etal-2020-state}
Pratik Joshi, Sebastin Santy, Amar Budhiraja, Kalika Bali, and Monojit Choudhury. 2020.
\newblock \href {https://doi.org/10.18653/v1/2020.acl-main.560} {The state and fate of linguistic diversity and inclusion in the {NLP} world}.
\newblock In \emph{Proceedings of the 58th Annual Meeting of the Association for Computational Linguistics}, pages 6282--6293, Online. Association for Computational Linguistics.

\bibitem[{Kamholz et~al.(2014)Kamholz, Pool, and Colowick}]{kamholz-etal-2014-panlex}
David Kamholz, Jonathan Pool, and Susan Colowick. 2014.
\newblock \href {http://www.lrec-conf.org/proceedings/lrec2014/pdf/1029_Paper.pdf} {{P}an{L}ex: Building a resource for panlingual lexical translation}.
\newblock In \emph{Proceedings of the Ninth International Conference on Language Resources and Evaluation ({LREC}'14)}, pages 3145--3150, Reykjavik, Iceland. European Language Resources Association (ELRA).

\bibitem[{Kargaran et~al.(2024)Kargaran, Yvon, and Sch{\"u}tze}]{kargaran-etal-2024-glotscript-resource}
Amir~Hossein Kargaran, Fran{\c{c}}ois Yvon, and Hinrich Sch{\"u}tze. 2024.
\newblock \href {https://aclanthology.org/2024.lrec-main.687} {{G}lot{S}cript: A resource and tool for low resource writing system identification}.
\newblock In \emph{Proceedings of the 2024 Joint International Conference on Computational Linguistics, Language Resources and Evaluation (LREC-COLING 2024)}, pages 7774--7784, Torino, Italia. ELRA and ICCL.

\bibitem[{Littell et~al.(2017)Littell, Mortensen, Lin, Kairis, Turner, and Levin}]{littell-etal-2017-uriel}
Patrick Littell, David~R. Mortensen, Ke~Lin, Katherine Kairis, Carlisle Turner, and Lori Levin. 2017.
\newblock \href {https://aclanthology.org/E17-2002} {{URIEL} and lang2vec: Representing languages as typological, geographical, and phylogenetic vectors}.
\newblock In \emph{Proceedings of the 15th Conference of the {E}uropean Chapter of the Association for Computational Linguistics: Volume 2, Short Papers}, pages 8--14, Valencia, Spain. Association for Computational Linguistics.

\bibitem[{Morey et~al.(2013)Morey, Post, and Friedman}]{morey2013language}
Stephen Morey, Mark~W Post, and Victor~A Friedman. 2013.
\newblock The language codes of iso 639: A premature, ultimately unobtainable, and possibly damaging standardization.

\bibitem[{Morgado~da Costa et~al.(2016)Morgado~da Costa, Bond, and Kratochv{\'\i}l}]{MorgadoDaCosta:Bond:Kratochvil:2016}
Luis Morgado~da Costa, Francis Bond, and Franti{\v{s}}ek Kratochv{\'\i}l. 2016.
\newblock Linking and disambiguating swadesh lists: Expanding the {Open Multilingual Wordnet} using open language resources.
\newblock In \emph{Proceedings of GLOBALEX 2016 Lexicographic Resources for Human Language Technology, 10th edition of the International Conference on Language Resources and Evaluation (LREC 2016)}, pages 29--36.

\bibitem[{Moseley(2010)}]{moseley2010atlas}
Christopher Moseley. 2010.
\newblock \emph{Atlas of the World's Languages in Danger}.
\newblock Unesco.

\bibitem[{of~Hawaii~at Manoa(2024)}]{elcat}
University of~Hawaii~at Manoa. 2024.
\newblock \href {http://www.endangeredlanguages.com} {Catalogue of endangered languages}.

\bibitem[{Swadesh(1955)}]{swadesh1955towards}
Morris Swadesh. 1955.
\newblock Towards greater accuracy in lexicostatistic dating.
\newblock \emph{International journal of American linguistics}, 21(2):121--137.

\bibitem[{Toossi et~al.(2024)Toossi, Huai, Liu, Khiu, Do{\u{g}}ru{\"o}z, and Lee}]{toossi2024reproducibility}
Hasti Toossi, Guo~Qing Huai, Jinyu Liu, Eric Khiu, A~Seza Do{\u{g}}ru{\"o}z, and En-Shiun~Annie Lee. 2024.
\newblock A reproducibility study on quantifying language similarity: The impact of missing values in the uriel knowledge base.
\newblock \emph{arXiv preprint arXiv:2405.11125}.

\bibitem[{{\"U}st{\"u}n et~al.(2022){\"U}st{\"u}n, Bisazza, Bouma, and van Noord}]{ustun-etal-2022-udapter}
Ahmet {\"U}st{\"u}n, Arianna Bisazza, Gosse Bouma, and Gertjan van Noord. 2022.
\newblock \href {https://doi.org/10.1162/coli_a_00443} {{UD}apter: Typology-based language adapters for multilingual dependency parsing and sequence labeling}.
\newblock \emph{Computational Linguistics}, 48(3):555--592.

\bibitem[{van~der Goot et~al.(2021)van~der Goot, {\"U}st{\"u}n, Ramponi, Sharaf, and Plank}]{van-der-goot-etal-2021-massive}
Rob van~der Goot, Ahmet {\"U}st{\"u}n, Alan Ramponi, Ibrahim Sharaf, and Barbara Plank. 2021.
\newblock \href {https://doi.org/10.18653/v1/2021.eacl-demos.22} {Massive choice, ample tasks ({M}a{C}h{A}mp): A toolkit for multi-task learning in {NLP}}.
\newblock In \emph{Proceedings of the 16th Conference of the European Chapter of the Association for Computational Linguistics: System Demonstrations}, pages 176--197, Online. Association for Computational Linguistics.

\bibitem[{Vastl et~al.(2020)Vastl, Zeman, and Rosa}]{vastl-etal-2020-predicting}
Martin Vastl, Daniel Zeman, and Rudolf Rosa. 2020.
\newblock \href {https://doi.org/10.18653/v1/2020.sigtyp-1.4} {Predicting typological features in {WALS} using language embeddings and conditional probabilities: {{\'U}FAL} submission to the {SIGTYP} 2020 shared task}.
\newblock In \emph{Proceedings of the Second Workshop on Computational Research in Linguistic Typology}, pages 29--35, Online. Association for Computational Linguistics.

\bibitem[{Wichmann et~al.(2022)Wichmann, Søren, Holman, and Brown}]{ASJP}
Wichmann, Søren, Eric~W. Holman, and Cecil~H. Brown. 2022.
\newblock The {ASJP} database (version 20).

\bibitem[{Yamada et~al.(2020)Yamada, Asai, Shindo, Takeda, and Matsumoto}]{yamada-etal-2020-luke}
Ikuya Yamada, Akari Asai, Hiroyuki Shindo, Hideaki Takeda, and Yuji Matsumoto. 2020.
\newblock \href {https://doi.org/10.18653/v1/2020.emnlp-main.523} {{LUKE}: Deep contextualized entity representations with entity-aware self-attention}.
\newblock In \emph{Proceedings of the 2020 Conference on Empirical Methods in Natural Language Processing (EMNLP)}, pages 6442--6454, Online. Association for Computational Linguistics.

\bibitem[{Yu et~al.(2021)Yu, He, and Sagae}]{yu-etal-2021-language}
Dian Yu, Taiqi He, and Kenji Sagae. 2021.
\newblock \href {https://aclanthology.org/2021.acl-long.560} {Language embeddings for typology and cross-lingual transfer learning}.
\newblock In \emph{Proceedings of the 59th Annual Meeting of the Association for Computational Linguistics and the 11th International Joint Conference on Natural Language Processing (Volume 1: Long Papers)}, pages 7210--7225, Online. Association for Computational Linguistics.

\end{thebibliography}
\clearpage
\appendix

\section{Hyperameters of feature presence classifier}
\label{app:finding-tuning}

\begin{figure}
    \includegraphics[width=\columnwidth]{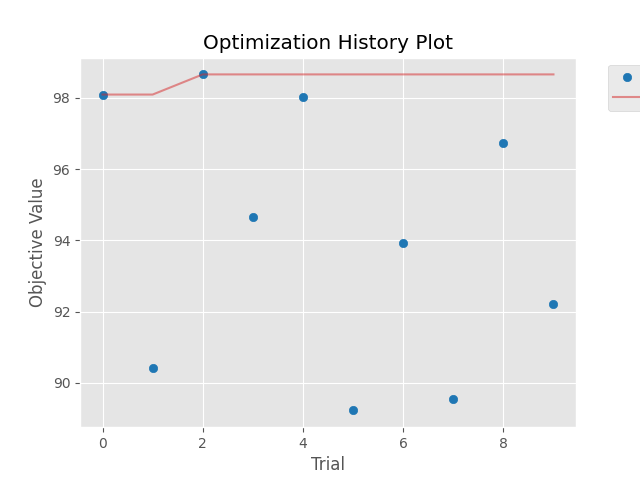}
    \caption{Optimization history for 10 trials, with the objective value being the F1-score for GBC Classifier.}
    \label{fig:rf_optimization_history}
\end{figure}

We use Optuna tools to select the suitable features and the best hyperparameters for the model for hyperparameter tuning and feature selection for missing values detection. Since hyperparameter tuning is costly, we randomly chose 300 languages from our dataset, extracting 39300 data points from these 300 languages. We use 5-fold cross-validation to validate the results. Figure \ref{fig:rf_optimization_history} shows the history of features and hyperparameter tuning.

Additionally, you can see which features were selected (true means the feature is selected, and false means the feature is ignored) and the best hyperparameter values used in 10 trials.

\begin{figure}
    \includegraphics[width=\columnwidth]{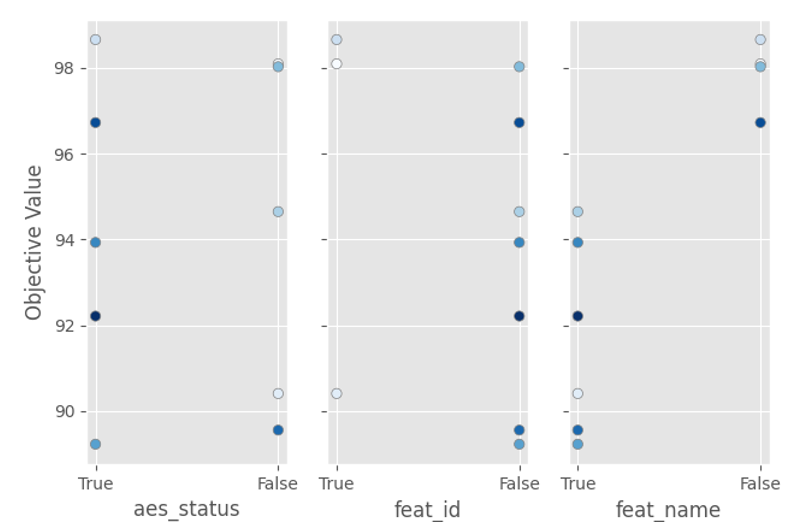}
    \caption{Optimization history for 10 trials in the GBC classifier for the \texttt{aes\_status}, \texttt{feat\_id}, and \texttt{feat\_name} features.}
    \label{fig:slice1}
\end{figure}

\begin{figure}
    \includegraphics[width=\columnwidth]{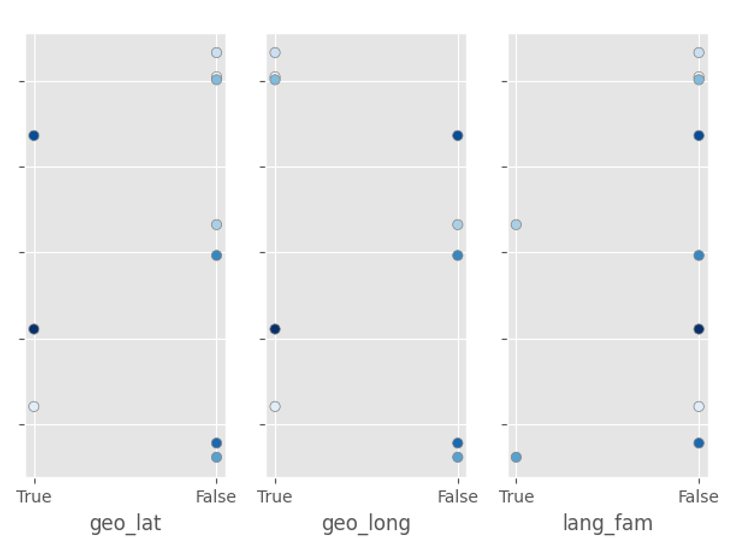}
    \caption{Optimization history for 10 trials in the GBC classifier for the \texttt{geo\_lat}, \texttt{geo\_long}, and \texttt{lang\_fam} features.}
    \label{fig:slice2}
\end{figure}

\begin{figure}
    \includegraphics[width=\columnwidth]{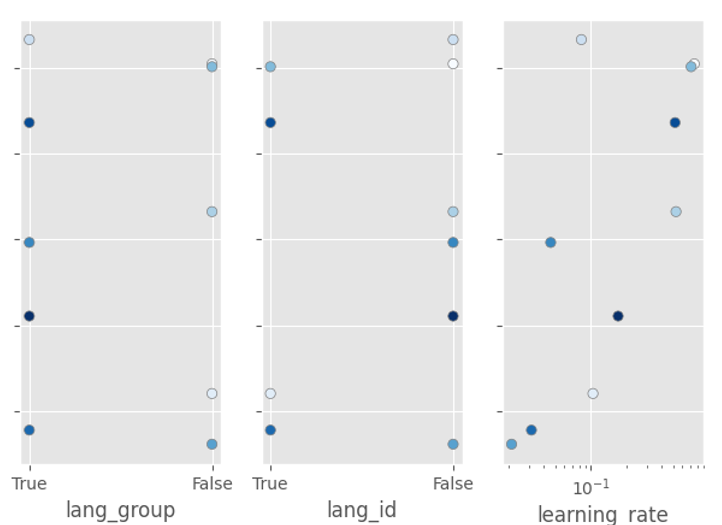}
    \caption{Optimization history for 10 trials in the GBC classifier for the \texttt{lang\_group}, \texttt{lang\_id}, and \texttt{learning\_rate} features.}
    \label{fig:slice3}
\end{figure}

\begin{figure}
    \includegraphics[width=\columnwidth]{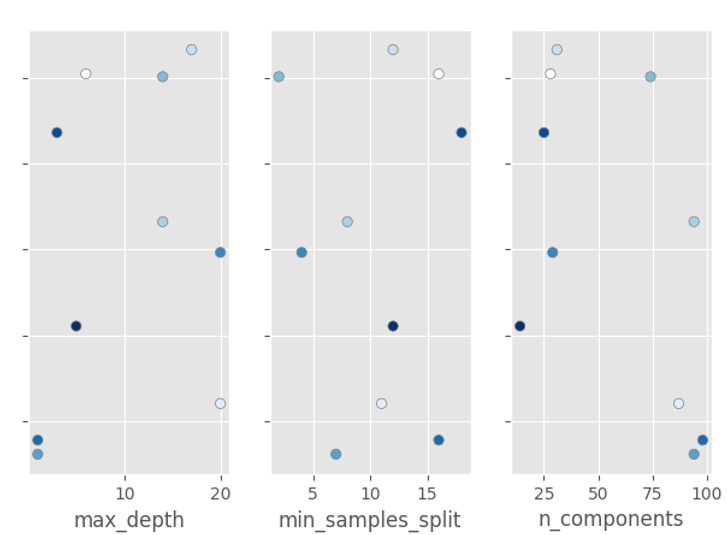}
    \caption{Optimization history for 10 trials in the GBC classifier for the \texttt{max\_depth}, \texttt{min\_samples\_split}, and \texttt{n\_components} features.}
    \label{fig:slice4}
\end{figure}

\begin{figure}
    \includegraphics[width=\columnwidth]{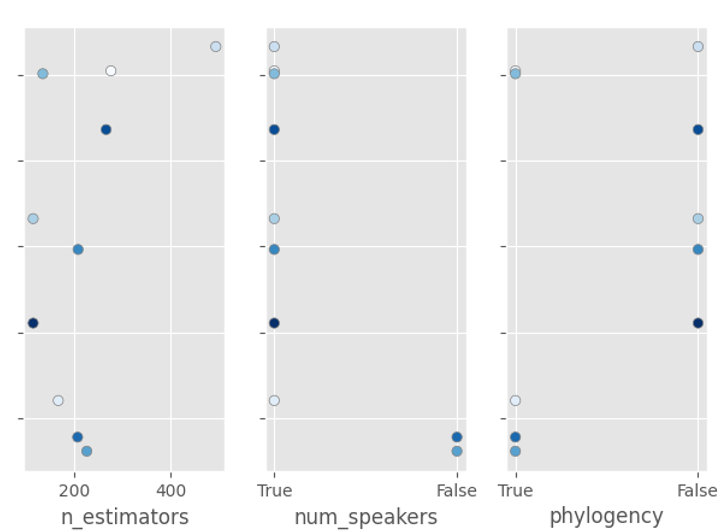}
    \caption{Optimization history for 10 trials in the GBC classifier for the hyperparameter \texttt{n\_estimators} and the features \texttt{num\_speakers} and \texttt{phylogency}.}
    \label{fig:slice5}
\end{figure}

\begin{figure}
    \includegraphics[width=\columnwidth]{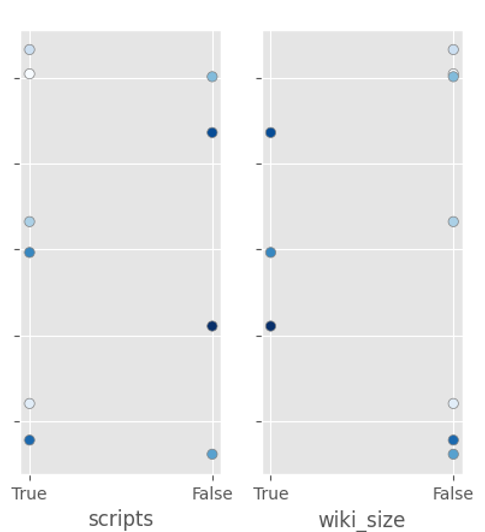}
    \caption{scripts, wiki\_size}
    \label{fig:slice6}
\end{figure}

\begin{table*}
    \centering
    \footnotesize
    \setlength{\tabcolsep}{3pt} 
    \begin{tabular}{l S[table-format=3.0] S[table-format=2.2] *{15}{c}} 
        \toprule
        \multirow{2}{*}{\textbf{Classifier}} & 
        \multicolumn{1}{c}{\textbf{\# Iterations}} & 
        \multicolumn{1}{c}{\textbf{F1 Score (\%)}} & 
        \multicolumn{15}{c}{\textbf{Feature Selection}} \\
        \cmidrule(lr){2-2} \cmidrule(lr){3-3} \cmidrule(l){4-18}
        & {} & {} & 
        \rotatebox{90}{\textbf{lang\_id}} & 
        \rotatebox{90}{\textbf{feat\_id}} & 
        \rotatebox{90}{\textbf{geo\_lat}} & 
        \rotatebox{90}{\textbf{geo\_long}} & 
        \rotatebox{90}{\textbf{lang\_group}} & 
        \rotatebox{90}{\textbf{aes\_status}} & 
        \rotatebox{90}{\textbf{wiki\_size}} & 
        \rotatebox{90}{\textbf{num\_speakers}} & 
        \rotatebox{90}{\textbf{lang\_fam}} & 
        \rotatebox{90}{\textbf{scripts}} & 
        \rotatebox{90}{\textbf{feat\_name}} & 
        \rotatebox{90}{\textbf{phylogeny}} & 
        \rotatebox{90}{\textbf{phylo\_n\_comp}} & 
        \rotatebox{90}{\textbf{miltale}} & 
        \rotatebox{90}{\textbf{miltate\_n\_comp}} \\
        \midrule
        Random Forest & 100 & 96.49 & $\times$ & $\times$ & \checkmark & $\times$ & $\times$ & $\times$ & $\times$ & \checkmark & $\times$ & $\times$ & \checkmark & \checkmark & 32 & $\times$ & $\times$ \\
        \midrule
        Logistic Regression & 100 & 95.39 & \checkmark & $\times$ & \checkmark & \checkmark & \checkmark & $\times$ & $\times$ & $\times$ & $\times$ & \checkmark & \checkmark & \checkmark & 18 & $\times$ & $\times$ \\
        \midrule
        K-Nearest Neighbor & 100 & 95.91 & $\times$ & \checkmark & $\times$ & $\times$ & \checkmark & $\times$ & \checkmark & $\times$ & \checkmark & \checkmark & \checkmark & \checkmark & 64 & $\times$ & $\times$ \\
        \midrule
        Gradient Boosting & 10 & \textbf{98.66} & \checkmark & $\times$ & \checkmark & $\times$ & $\times$ & $\times$ & \checkmark & $\times$ & \checkmark & $\times$ & \checkmark & \checkmark & 31 & $\times$ & $\times$ \\
        \midrule
        Decision Tree & 100 & 97.56 & \checkmark & $\times$ & $\times$ & $\times$ & \checkmark & \checkmark & \checkmark & \checkmark & $\times$ & $\times$ & \checkmark & $\times$ & $\times$ & $\times$ & $\times$ \\
        \bottomrule
    \end{tabular}
    \caption{K-fold cross-validation results for feature presence classifier.}
    \label{tab:feature_selection}
\end{table*}

The best hyperparameters for the Gradient Boosting classifier are as follows: a maximum depth of 17, a minimum sample split of 12, a learning rate of 0.0836, 494 estimators, and 31 components for the phylogeny PCA dimensions. These settings have been found to optimize the classifier's performance.

\section{Setup}
\begin{figure}
\includegraphics[width=\columnwidth]{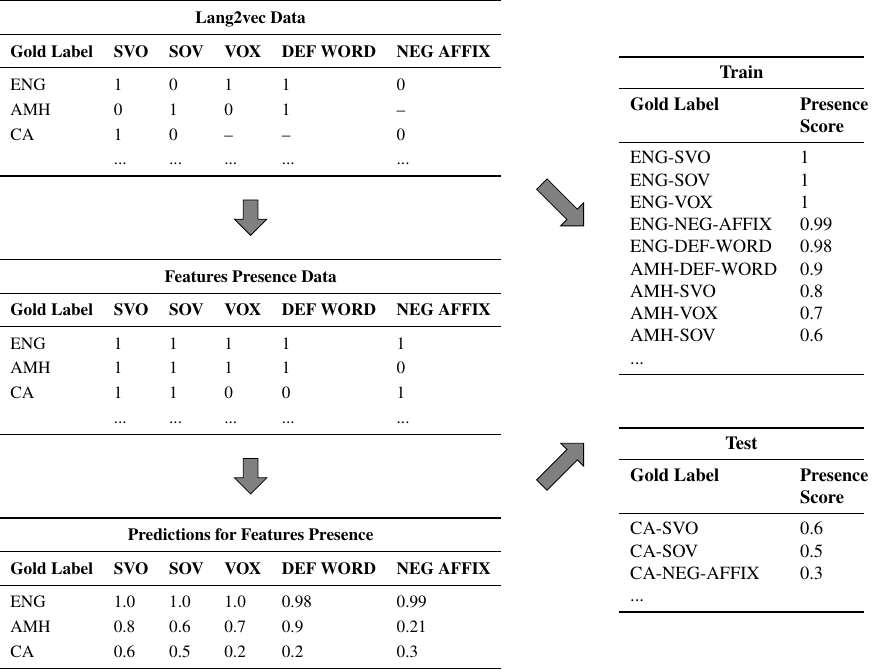}
\caption{Schematic overview of the setup, where we first identify likely missing values to create our final evaluation setup.}
\label{fig:setup}
\end{figure}

An overview of our setup is shown in Figure~\ref{fig:setup}.
\label{app:setup}

\section{Results of POS taggers}
\label{app:posresults}
\begin{figure}
    \includegraphics[width=\columnwidth]{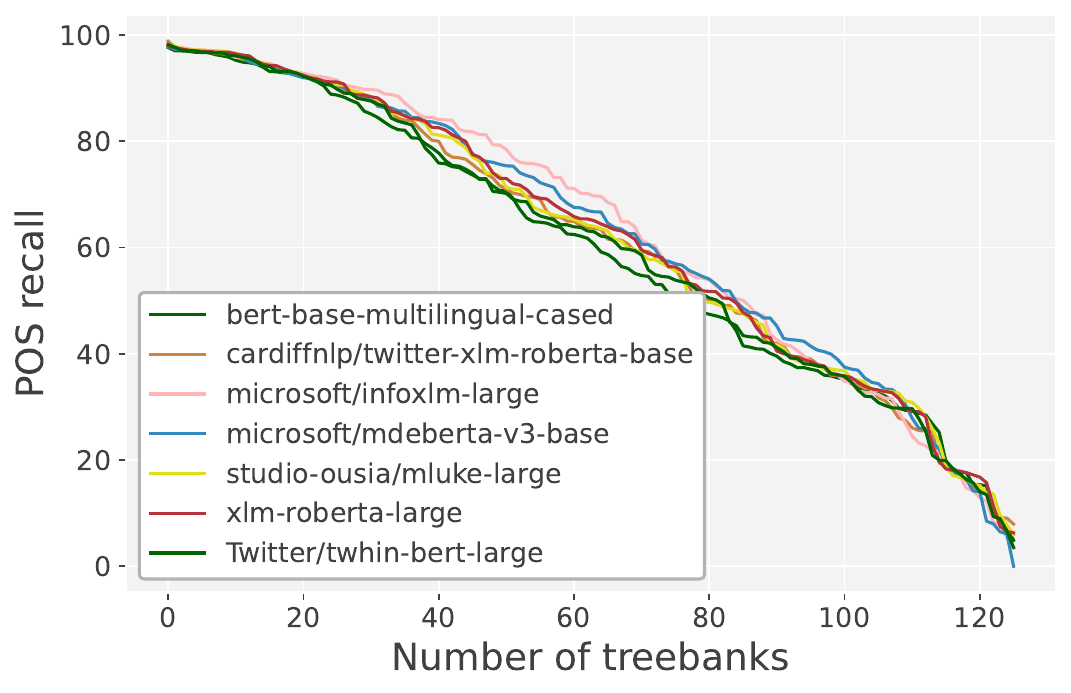}
    \caption{\% Recall for POS tagging of treebanks that were not included in training}
    \label{fig:newtreebanks}
\end{figure}

\begin{figure}[t]
    \includegraphics[width=\columnwidth]{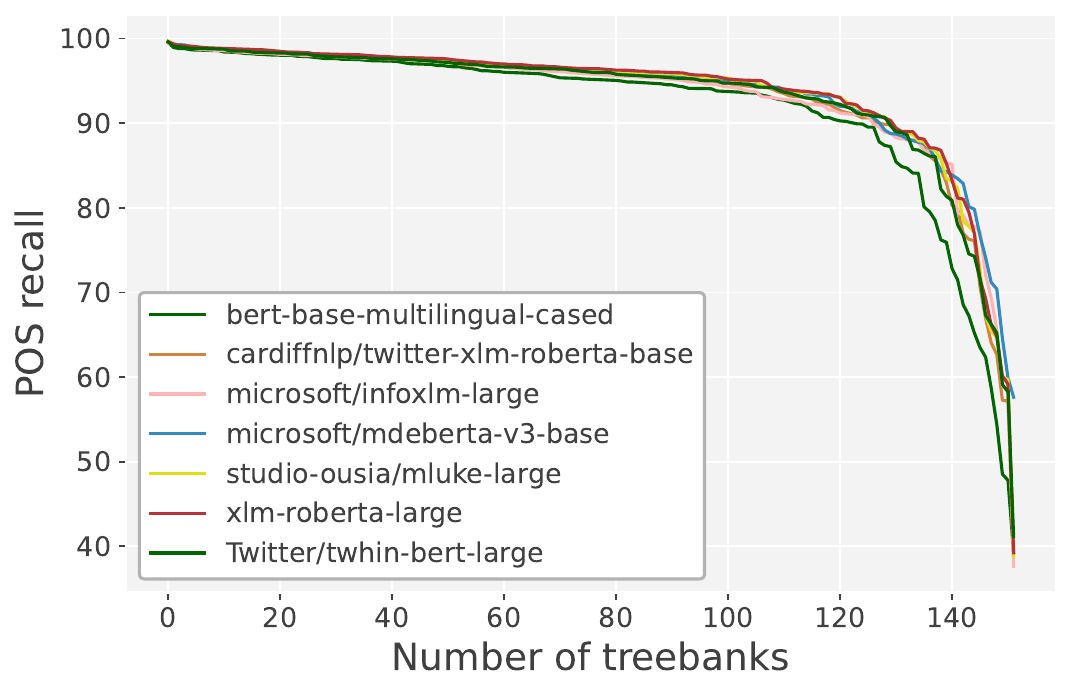}
    \caption{\% Recall for POS tagging of test-splits of treebanks that the taggers were trained on.}
    \label{fig:seentreebanks}
\end{figure}

Figure~\ref{fig:newtreebanks} and Figure~\ref{fig:seentreebanks} show the performance of the POS taggers as recall. The average scores for each language model are reported in Table~\ref{tab:avgpos}. We based our selection of the language model on the performance on the 
new (i.e. not included in training) treebanks, as they are most similar to the majority of the languages in our data.

To be able to do filtering based on POS tagging accuracy, we need to be able to predict the performance of the POS tagger for unlabeled instances (i.e. our plain text dataset). For this, we train machine learning classifiers with the following features:
\begin{itemize}
    \item Frequency as the probability for each POS tag
    \item Average confidence (logit after softmax) over all data from the language
    \item Percentage of UNK subwords
    \item Average length of words in \#subwords
    \item Average length of words in \#characters
    \item Percentage of correct POS tagging labels for the language in Swadesh lists~\cite{swadesh1955towards}, aligned with the English POS tags from the same tagger. We combined Swadesh lists from PanLex~\cite{kamholz-etal-2014-panlex} and~\newcite{MorgadoDaCosta:Bond:Kratochvil:2016} for greater coverage.
\end{itemize}

Based on these features, we evaluate a random forest classifier, an SVR classifier, a Lasso classifier, and an elastic search. We achieve the best performance with a random forest classifier with 200 estimators, with an average distance to the actual performance of 7.28. Although this might seem like a high number, we only use this to differentiate the data roughly in 2 parts: one where the POS tagger has learned some notion of the task for the target language and one where performance is so low that it is completely unusable.

Based on these scores, we have 1,749 languages with an estimated performance $>$ 70, and 559 with a score $>$ 80. For our studies, we focus on a more accurate sample of 559 languages.

\begin{table}[t]
\resizebox{1\columnwidth}{!}{
\begin{tabular}{l r r}
\toprule
LM & avg. new & avg. seen \\
\midrule
bert-base-multilingual-cased & 59.82 & 91.67\\  
cardiffnlp/twitter-xlm-roberta-base & 61.34 & 93.15\\
microsoft/infoxlm-large & 63.35 & 93.21\\
microsoft/mdeberta-v3-base & 62.88 & 93.91\\
studio-ousia/mluke-large & 61.86 & 93.64\\
xlm-roberta-large & 62.24 & 93.79\\
Twitter/twhin-bert-large & 60.59 & 93.22\\
\bottomrule
\end{tabular}}
\caption{Average \% recall for each language model we evaluated.}
\label{tab:avgpos}
\end{table}

\end{document}